\documentclass[letterpaper, 10 pt, journal, twoside]{IEEEtran}
\usepackage{amsmath,amsfonts}
\usepackage{algorithmic}
\usepackage{algorithm}
\usepackage{array}
\usepackage[caption=false,font=normalsize,labelfont=sf,textfont=sf]{subfig}
\usepackage{textcomp}
\usepackage{stfloats}
\usepackage{url}
\usepackage{verbatim}
\usepackage{graphicx}
\usepackage{cite}
\usepackage{bm}
\usepackage{times}
\usepackage{epsfig}
\usepackage{amssymb}
\usepackage{multirow}
\usepackage{multicol}
\begin{document}

\title{DVGG: Deep Variational Grasp Generation for Dextrous Manipulation}

\author{Wei Wei, Daheng Li, Peng Wang, \textit{Member, IEEE}, Yiming Li, Wanyi Li, Yongkang Luo, Jun Zhong
\thanks{
This work was supported in part by the InnoHK, in part by the National Natural Science Foundation of China under Grants (91748131, 62006229 and 61771471), and in part by the Strategic Priority Research Program of Chinese Academy of Science under Grant XDB32050106. (\textit{Corresponding author: Peng Wang}.)}
\thanks{Peng Wang is with Institute of Automation, Chinese Academy of Sciences, Beijing 100190, China, and with the School of Artificial Intelligence, University of Chinese Academy of Sciences, Beijing 100049, China, and with the CAS Center for Excellence in Brain Science and Intelligence Technology, Chinese Academy of Sciences, Shanghai 200031, China, and also with the Centre for Artificial Intelligence and Robotics, Hong Kong Institute of Science and Innovation, Chinese Academy of Sciences, Hong Kong 999077, China (email: peng\_wang@ia.ac.cn).}
\thanks{Wei Wei, Daheng Li, Yiming Li are with Institute of Automation, Chinese Academy of Sciences, Beijing 100190, China, and with the School of Artificial Intelligence, University of Chinese Academy of Sciences, Beijing 100049, China.}
\thanks{Wanyi Li, Yongkang Luo, Jun Zhong are with Institute of Automation, Chinese Academy of Sciences, Beijing 100190, China.}
}

\maketitle

\begin{abstract}
Grasping with anthropomorphic robotic hands involves much more hand-object interactions compared to parallel-jaw grippers. Modeling hand-object interactions is essential to the study of multi-finger hand dextrous manipulation. This work presents DVGG, an efficient grasp generation network that takes single-view observation as input and predicts high-quality grasp configurations for unknown objects. In general, our generative model consists of three components: 1) Point cloud completion for the target object based on the partial observation; 2) Diverse sets of grasps generation given the complete point cloud; 3) Iterative grasp pose refinement for physically plausible grasp optimization. To train our model, we build a large-scale grasping dataset that contains about 300 common object models with 1.5M annotated grasps in simulation. Experiments in simulation show that our model can predict robust grasp poses with a wide variety and high success rate. Real robot platform experiments demonstrate that the model trained on our dataset performs well in the real world. Remarkably, our method achieves a grasp success rate of 70.7\% for novel objects in the real robot platform, which is a significant improvement over the baseline methods.
\end{abstract}


\section{Introduction}
\IEEEPARstart{G}{rasping} with parallel-jaw grippers has been well investigated and widely applied in robotic manipulation. However, the study of anthropomorphic robotic hand remains a challenge for the robotics community. Multi-finger grippers equipped with multiple actuated joints enable robots to perform more advanced operations such as grasping for tool-use.

Traditional analysis-based methods for multi-finger grasp generation rely on the assumption that object shapes and poses are known a priori \cite{graspit,eigengrasp, dai2018synthesis} that they can synthesize grasp configurations with commonly used grasping metrics, \textit{e.g.} force-closure \cite{force_closure} and $\epsilon$-metric \cite{ferri_canny}. However, these approaches are not applicable to unseen objects. To generalize to unknown objects, most current works utilize shape completion module \cite{varley2017shape,lundell2019shape}, while these methods are time-consuming due to the huge search space for high-DOF grippers.
\begin{figure}
    \centering
    \includegraphics[width=0.85\linewidth]{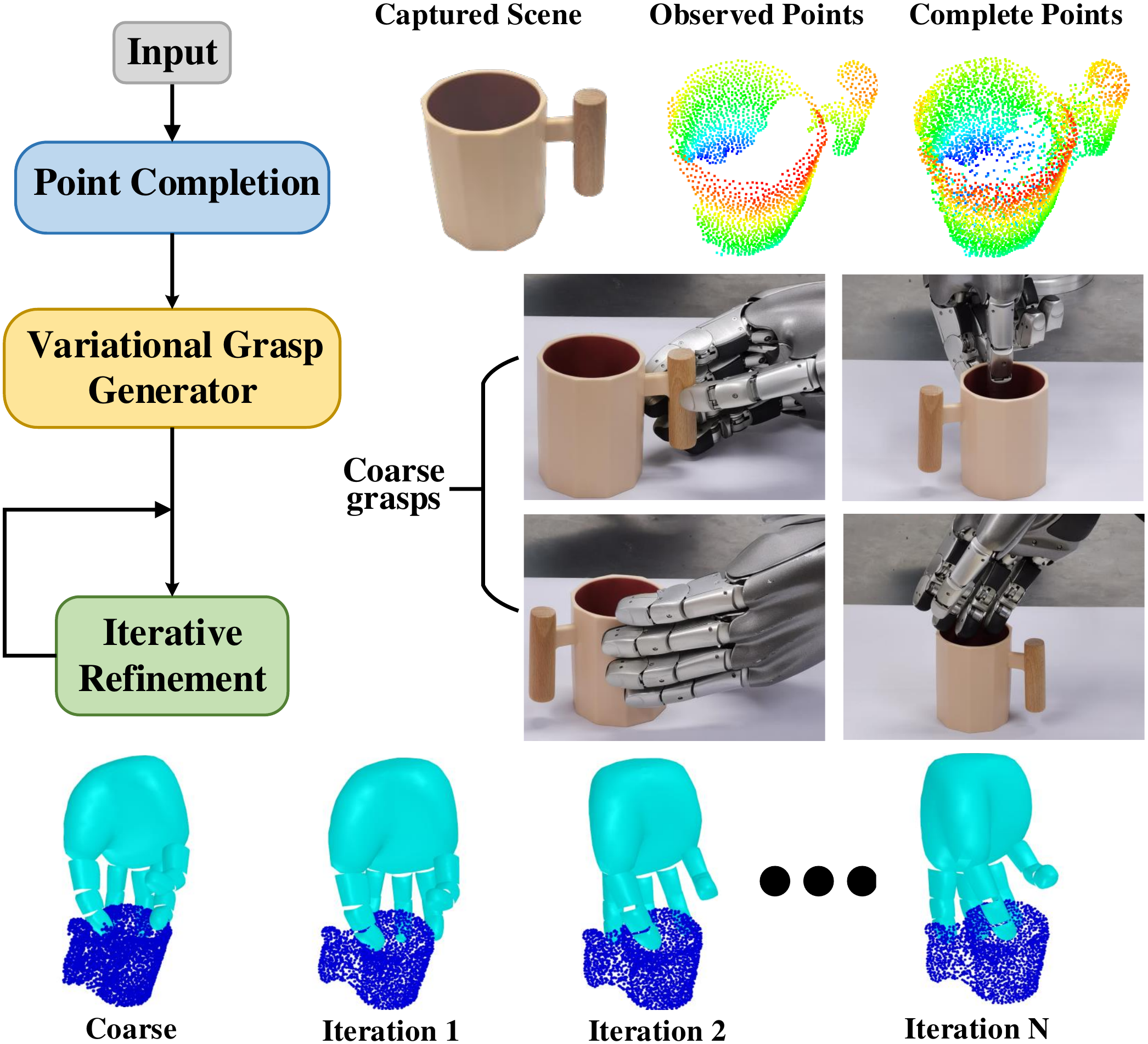}
    \vspace{-3mm}
    \caption{Grasping a target object with dextrous grasp configurations. Our approach is able to generate diverse sets of grasps for unknown objects efficiently in a coarse to fine manner.}
    \label{fig:idea}
    \vspace{-7mm}
\end{figure}

In recent years, learning-based approaches have made significant breakthroughs. Most learning-based grasping methods estimate grasp poses directly from raw sensor inputs \cite{wei2021gpr,image_grasp3,rgbd_matter,6dofgraspnet}. Nevertheless, most of these approaches focus on studying parallel-jaw grippers with the majority of degrees of freedom concentrated in the wrist joint.

Although it is challenging to predict high-DOF grasp poses, researchers have proposed various solutions to this problem. \cite{varley2015generating} uses a neural network to predict pixel-wise heatmap for multi-finger placement but relies on a grasp planner to determine the final grasp pose. \cite{DDG,kopicki2019learning} train evaluation models with grasps generated by a grasp sampler, which involves manually-engineered mappings from observation to grasp. \cite{highdofgrasp, deep_differentiable_grasp_planner} propose to identify a one-to-one mapping from objects to grasp poses, which results in limited grasp postures. \cite{MDN,lu2020multi} learn a grasp-success probability prediction model with a voxel-based 3D convolutional neural network, while the hand needs to approach the object with limited directions. \cite{wu2020generative} proposes a ``GenerAL" framework with reinforcement learning for multi-fingered grasping in clutter, while encountering the sim-real gap problem.

Meanwhile, a large body of works concentrated on human grasp estimation \cite{jiang2021hand,grabnet,ganhand,contactpose} achieve promising results on daily objects \cite{ycb}. These works focus on hand-object interactions based on contacts, which is intuitive for human grasping.

Inspired by this intuitive idea, we propose to generate diverse sets of robotics grasps based on contacts. As shown in Fig.\ref{fig:idea}, our method works in the following way: 1) Complete points are first reconstructed by the point completion module; 2) The variational grasp generator estimates diverse sets of coarse grasps given the complete object points; 3) Coarse grasps are further refined by the iterative refinement module.

We evaluate our approach on the YCB \cite{ycb} and EGAD! \cite{egad} dataset. Experimental results show that our model can generate diverse sets of grasps in terms of approach direction and posture, and the refinement module helps to produce physically plausible grasps. Furthermore, our method shows significant improvement compared to baselines in terms of time efficiency and grasp success rate.

In summary, our primary contributions are: 1) A novel generative model for generating diverse sets of high-DOF hand grasps based on hand-object interactions; 2) A useful iterative refinement module for hand grasp refinement; 3) Significant improvement on grasp generation for both seen and unseen objects compared to the baseline methods.


\section{RELATED WORK}
\textbf{Grasp Planners.} Traditional grasp planners \cite{graspit, fast_planner,eigengrasp,force_closure} assumed that the environment is fully perceptible. Various algorithms \cite{dai2018synthesis,eigengrasp,force_closure} in this direction have been proposed to search for high-quality grasps based on certain grasp metrics \cite{ferri_canny,zheng2012efficient}. However, these methods are challenging for a number of factors: 1) Environments are not fully perceptible, especially for unstructured environments; 2) Time-consuming due to the large searching space for high-DOF grippers. Recently, learning-based grasp planners propose to predict grasps directly given the partial observation. Most works following this way \cite{dexnet2.0,wei2021gpr,image_grasp2, image_grasp3} focus on studying parallel-jaw grippers with different kinds of inputs, \textit{e.g.} RGB-D images or point clouds. \cite{image_grasp2,image_grasp3} propose to conduct grasp pose detection as rectangle detection in 2D space, the grippers are limited to approaching the target object vertically. DexNet \cite{dexnet2.0} collects numerous object models for GQ-CNN training and achieves state-of-the-art performance. \cite{wei2021gpr,fang2020graspnet} propose to estimate 6 or 7 DOF grasp poses in cluttered scenes. However, for high-DOF anthropomorphic hands, in addition to the wrist pose, remaining hand joints play a more important role in human-like manipulation such as in-hand manipulation. 

\textbf{Learning for hand-object interactions.} The research of hand-object interactions has been widely studied in the computer vision community \cite{ganhand,grabnet,jiang2021hand}. Recently, several datasets are proposed for facilitating hand-object interactions research. \cite{ho3d,grabnet,fhab} label human hand grasps with captured images or videos.  \cite{ganhand,hasson2019learning} propose to synthesize human hand grasps with GraspIt!\cite{graspit}, while the grasps may not look realistic in general. Most of these methods propose to predict affordance map for target objects, and predict grasps based on contacts and penetration jointly. However, only a few researchers pay attention to hand-object contacts in robotics grasping \cite{multifingan,ddgc,contactgrasp,unigrasp}. To our knowledge, \mbox{\cite{contactgrasp}} is the most similar work to ours, which refines sampled grasps with human-demonstrated contact map for functional grasping. In this work, we propose to generate grasps with implicitly hand-object contacts based on a generative model.

\section{PROBLEM STATEMENT}
This work concentrates on planning dextrous high-DOF robotic hand grasping based on single view observation, which implies generating physically plausible and collisionless grasp configurations. More formally, our model $\mathcal{M}$ takes the observed point cloud $\mathcal{P}$ as input and predicts high quality grasps. Each grasp $\bm g$ is represented by a hand wrist pose $\mathbf{p}$ and a hand joint configuration $\bm{\theta}$, $\textit{i.e.}$ $\bm g = \left\{\mathbf{p}, \bm{\theta}\right\}$. Hand wrist pose $\mathbf{p}$ is given in $\mathcal{S}{E}(3)$, including the translation $\mathbf{t}=[t_x, t_y, t_z]$ and orientation quaternion $\mathbf{q}=[q_w, q_x, q_y, q_z]$. Hand joint configuration $\bm{{\theta}}$ is denoted by the actual degree of freedom of the hand,  $\bm{\theta} \in \mathbb{R}^{20}$ for HIT-DLR \uppercase\expandafter{\romannumeral2} Hand.


\section{DATASET GENERATION}
In this section, we give a brief introduction for the procedure of building our grasp dataset in the physics simulator.

\textbf{Objects.} We collect 300 objects from the dataset released in the previous work \cite{deep_differentiable_grasp_planner}. Some of the objects are re-scaled to fit for the HIT-DLR \uppercase\expandafter{\romannumeral2} Hand. All objects share the same coefficient of friction (0.25) and density (1500kg/$m^3$).

\begin{figure}
    \begin{center}
		\includegraphics[width=0.9\linewidth]{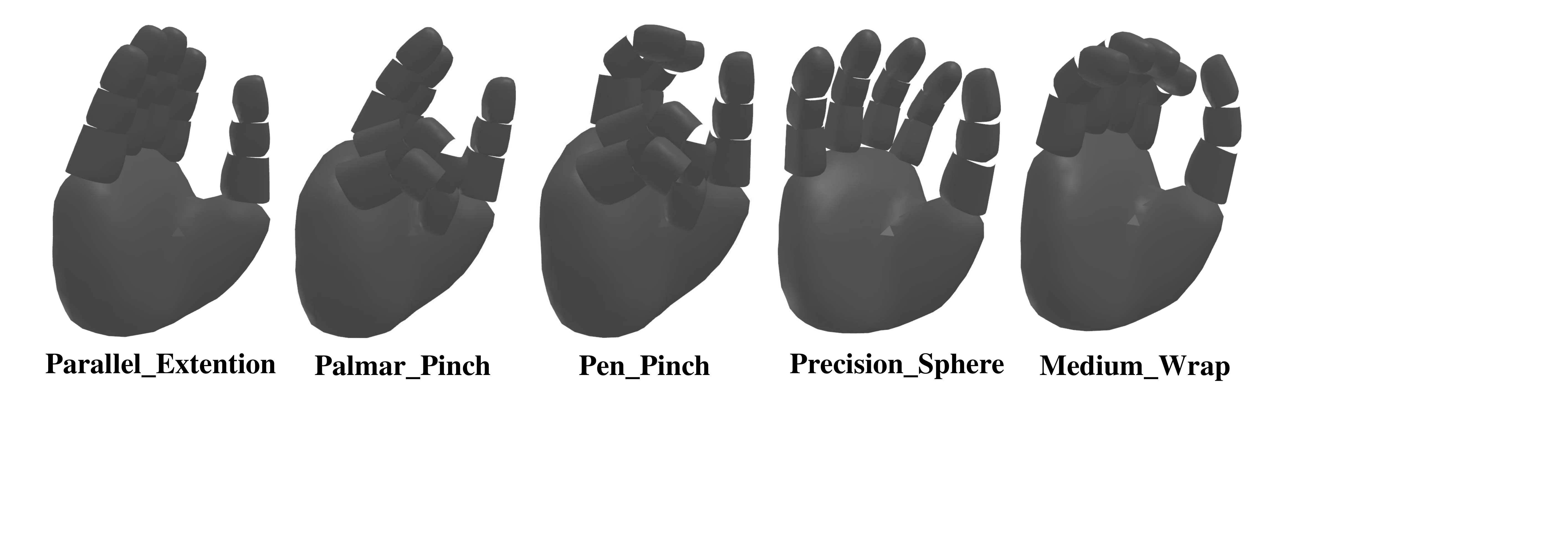}
	\end{center}
	\vspace{-3mm}
    \caption{Five taxonomy for annotating grasps in simulation.}
    \label{fig:five_taxonomy}
    \vspace{-6mm}
\end{figure}

\textbf{Grasp Annotation.} We adopt the commonly used approach direction sampling scheme. It first samples a point on the surface,  then the hand approaches the object and executes a grasp attempt with a predefined step interval. For each step interval, uniform in-plane rotations are sampled. Five taxonomy shown in Fig.\ref{fig:five_taxonomy} are selected for annotation\cite{feix2015grasp}.

\textbf{Grasp Labelling in Simulation.} We build our synthetic grasping dataset in the physics simulator MuJoCo \cite{mujoco}.  The physics simulation consists of three steps: 1) Objects are fixed stationary at first and the hand is  initialized  with a pre-grasp state, then the hand approaches the object and executes a  grasp attempt with a pre-defined grasp taxonomy until the simulator reaches a stable state. At this time, all fingers should contact the object or reach their maximum joint angles. 2) Then all fingers keep the grasping force while the gravity is present till the simulator reaches a stable state or the object falls from the hand. 3) Unstable grasps are filtered by shaking the hand and keep those grasps that consistently keep the object in hand.

\begin{figure*}
    \begin{center}
		\includegraphics[width=0.9\linewidth]{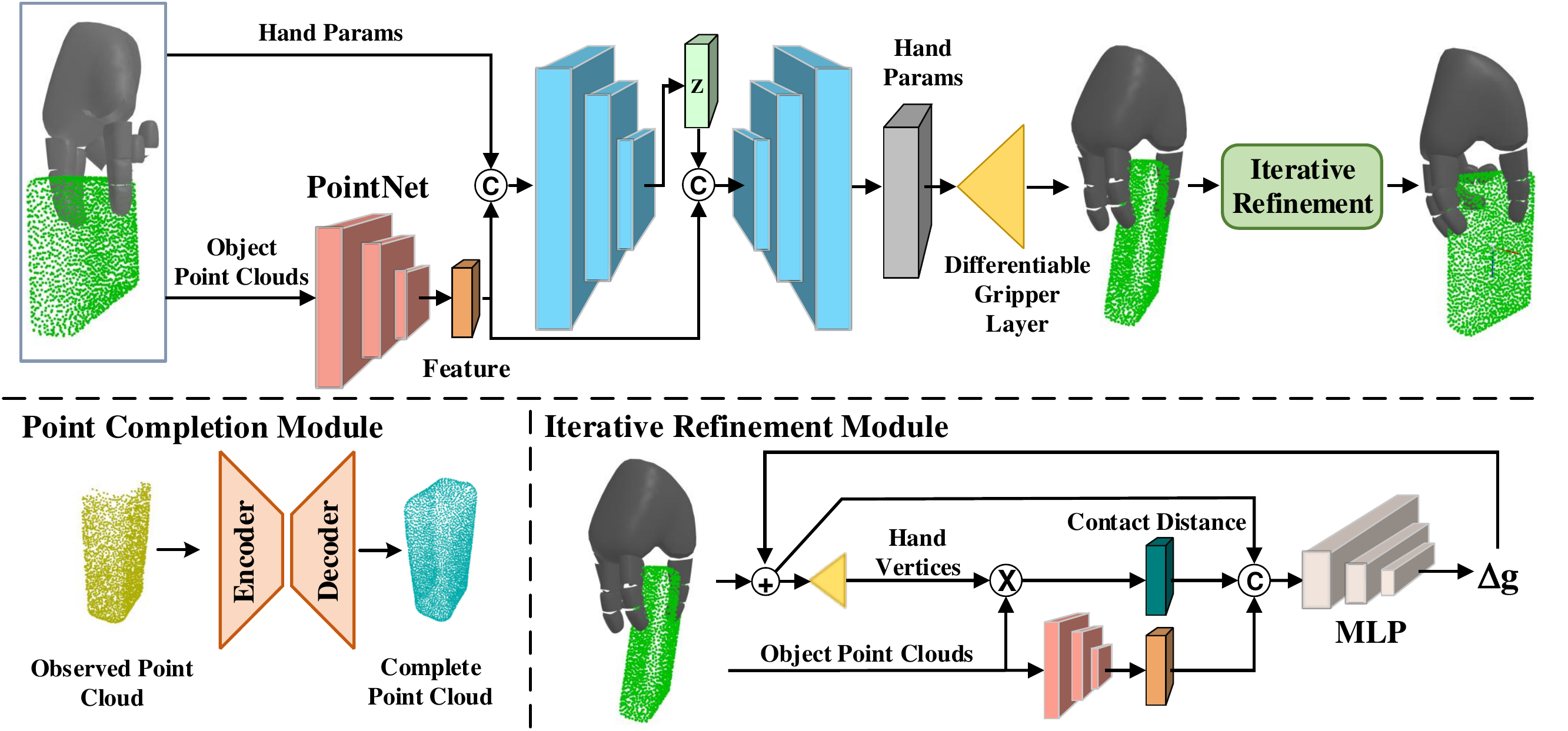}
	\end{center}
	\vspace{-6mm}
    \caption{Overview architecture of \textbf{DVGG} network for dextrous grasp pose generation and refinement. Point Completion Module for generating complete point clouds from partial object point clouds, shown in the left bottom. Iterative Refinement Module for grasp pose refinement, shown in the right bottom.}
    \label{fig:pipeline}
    \vspace{-6mm}
\end{figure*}

\section{METHOD}
In this section, we present the deep variational grasp generation network (\textbf{DVGG}) for generating dextrous grasp poses. The overall pipeline is shown in Fig.\ref{fig:pipeline}. It consists of 3 submodules: Object Point Completion, Variational Grasp Generator, and Iterative Grasp Refinement. 

\subsection{Object Point Completion}
Manipulation of anthropomorphic hands involves rich hand-object contacts. Most of the existing methods take the assumption that the target object model is known \mbox{\cite{jiang2021hand,grabnet,contactgrasp}}. Some of the current works propose to utilize multi-view inputs for either object reconstruction \mbox{\cite{highdofgrasp}} or feature aggregation \mbox{\cite{deep_differentiable_grasp_planner,contactpose}}, while multi-camera platform setup imposes strict restrictions for real-world applications.

\begin{figure}[t]
    \centering
    \includegraphics[width=0.85\linewidth]{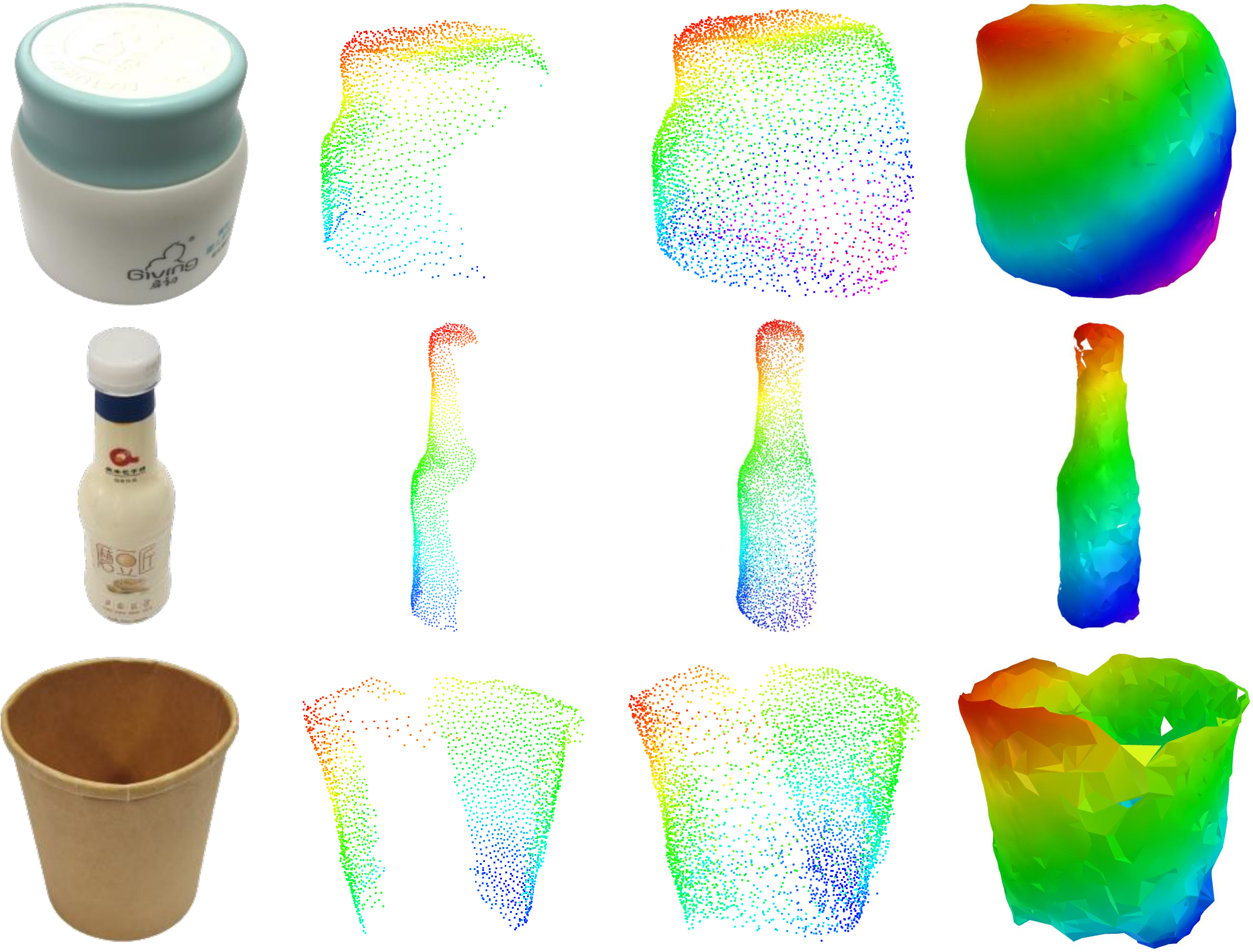}
    \vspace{-3mm}
    \caption{Qualitative results of Point Completion Module on unseen objects. The first column shows the real-world object, the second and the third show the partial observed and predicted complete point clouds, the last column shows the reconstructed surface mesh.}
    \label{fig:point_completion_examples}
\end{figure}

To this end, we propose to directly estimate the complete object model from single-view input inspired by \cite{multifingan}. In practice, we follow the method proposed by \cite{vrcnet} and make two modifications during data synthesis to make it easy for real platforms application: 1) Position of partial observed point clouds are represented in camera coordinate system instead of the object coordinate system; 2) Origin is normalized to the center of the partial observed point clouds instead of the centroid of the objects. We synthesize about 1M partial observed point clouds, and train the model from scratch. All objects come from \cite{deep_differentiable_grasp_planner}. Experimental results shown in Fig.\ref{fig:point_completion_examples} demonstrate that the model trained on our dataset performs well on real robotic platform.


\subsection{Variational Grasp Generator}
The Variational Grasp Generator is based on Conditional Variational Auto-Encoder (CVAE) \cite{vae} generative network. For training our grasp generator, we use ground-truth complete object point clouds as conditional information. In the training stage, both the encoder and decoder are used to learn the grasp generation task by optimizing the reconstruction errors with hand-object interactions constraint; At testing time, only the decoder is used. The network architecture is shown in Fig.\ref{fig:pipeline}. Both the encoder and decoder of the generator are composed of Multi-Layer Perceptrons (MLP).
 
During training stage, given the grasp configuration $\bm{g} \in \mathbb{R}^{27}$ and the object point cloud $\mathcal{P}^o \in \mathbb{R}^{N\times3}$ as input, we utilize PointNet \cite{pointnet} to extract point cloud feature $\mathcal{F}^o$ and MLPs for hand configuration $\bm g$ feature extraction  $\mathcal{F}^h$. These two features are then concatenated together as the input for the encoder. The encoder learn to map each pair of point cloud $\mathcal{P}^o$ and grasp $\bm g$ to a subspace in the latent space $\bm {z}$, where $P(\bm z) \sim \mathcal{N}(0, I)$. 

Given the sampled latent code $\bm z$ and the extracted feature $\mathcal{F}^o$ as input to the decoder, the decoder learns to predict the hand parameters $\bm{g}$. Given $\bm {g}$ as input, the hand mesh is reconstructed by a differentiable hand layer. This layer is designed based on the forward kinematics of the hand. It takes the hand parameters as input and outputs the mesh $\bm{\mathcal{H}}=(\bm {V}, \bm {F})$, where $\bm V$ and $\bm F$ denote vertices and faces.

During testing stage, the encoder is removed, and a latent value $\bm z$ is randomly sampled from Normal Gaussian Distribution $\mathcal{N}(0, I)$. Then the latent code $\bm {z}$ and the extracted object feature $\mathcal{F}^o$ are concatenated together as input to the decoder. The decoder predicts hand parameters for the differentiable hand layer to reconstruct the hand model.

Given the above introduction for network architecture and workflow of the grasp generator, we then present the detailed objective function for training our network. If not specified, a hat on top refers to predicted variables.

\textbf{KL-Divergence}. We use a KL-divergence loss to regularize the distribution of latent code $\bm{z}$, enforcing the latent code distribution $Q(\bm z|\mathcal{P}^o, \bm{g})$ to be close to a standard Gaussian Distribution. This can be done by maximizing the KL-Divergence as follows:
\begin{equation}
\begin{aligned}
     \mathcal{L}_{\mathcal{KL}} &= -{\bm{KL}}\left( Q(\bm z|\mathcal{P}^o, \bm{g}) || \mathcal{N}(0, I)\right)
\end{aligned}
\label{equation:kl_loss}
\end{equation}

\textbf{Reconstruction}. We design the reconstruction objective based on the reconstructed hand mesh. It consists of the following terms: hand mesh vertices displacement and joint angles $\bm{\theta}$ error. We adopt the $\bm{L}_2$ loss for optimizing Euclidean distance of the vertices and the $\bm{L}_1$ loss for the joint angles. The losses are formulated as follows:
\begin{equation}
\begin{aligned}
     \mathcal{L}_{\mathcal{V}} &= \frac{1}{N}\sum_{i}^{N}\bm{||} \hat{\bm{V}_i} - \bm{V}_i\bm{||}^2_2 \\
     \mathcal{L}_{\bm{\theta}} &= \bm{|} \hat{\bm{\theta}} - \bm{\theta}\bm{|} \\
     \mathcal{L}_{\mathcal{R}} &= \lambda_{\mathcal{V}} \cdot \mathcal{L}_{\mathcal{V}} + \lambda_{\theta} \cdot \mathcal{L}_{\bm{\theta}}
\end{aligned}
\label{equation:reconstruction_loss}
\end{equation}
The reconstruction loss $\mathcal{L}_{\mathcal{R}}$ include two terms, $\mathcal{L}_{\mathcal{V}}$ for hand mesh vertices reconstruction and $\mathcal{L}_{\bm{\theta}}$ for hand joint angles prediction. Where $N$ is the number of hand mesh vertices, $\bm{V}$ denotes the hand mesh vertices, $\bm{\theta}$ denote the hand joint angles. $\lambda_{\mathcal{V}}$ and $\lambda_{\theta}$ are constants for balancing the losses.

\begin{figure}
    \centering
    \subfloat[]{\includegraphics[height=0.4\linewidth]{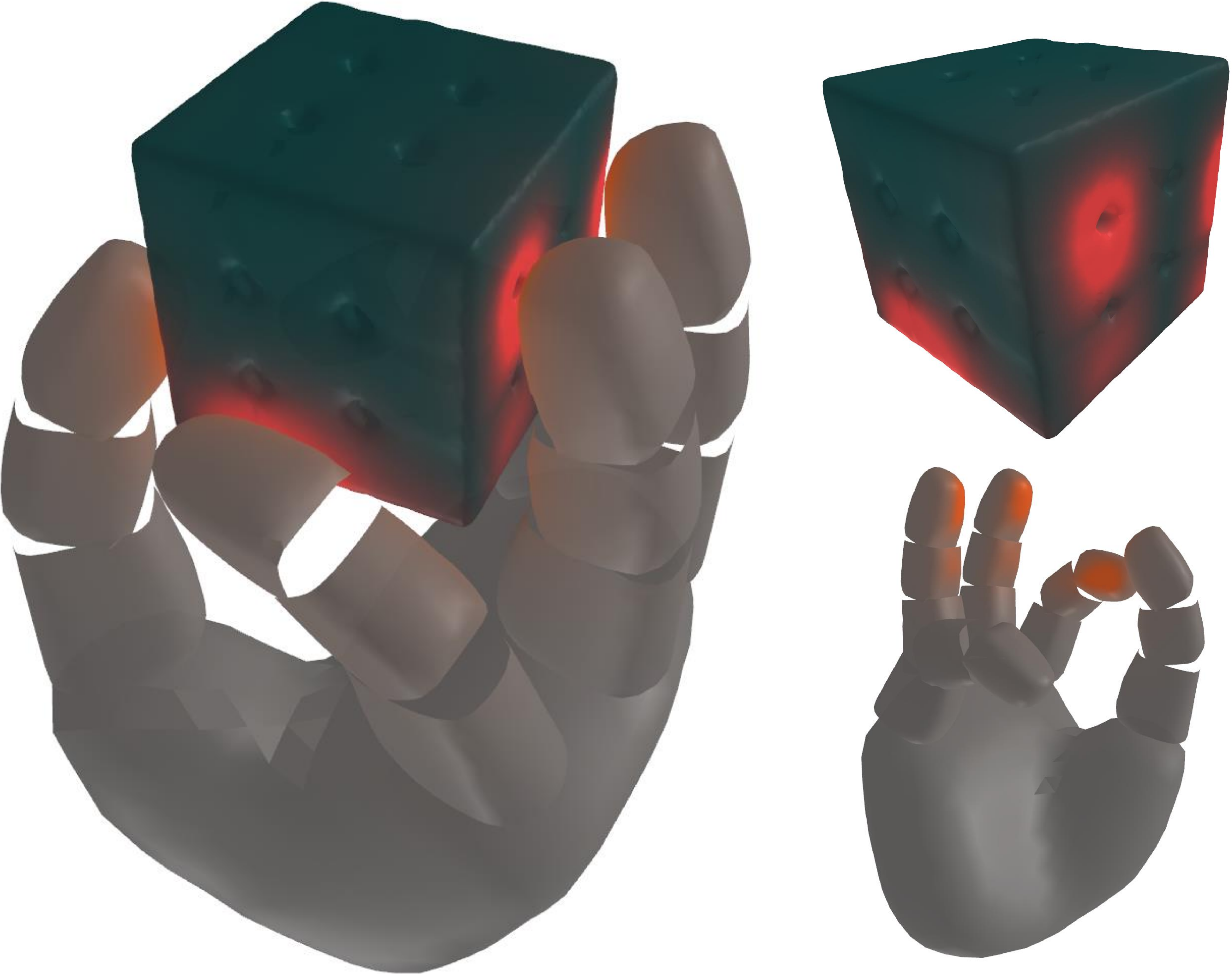}
        \label{fig:contact_map}}
    \subfloat[]{\includegraphics[height=0.4\linewidth]{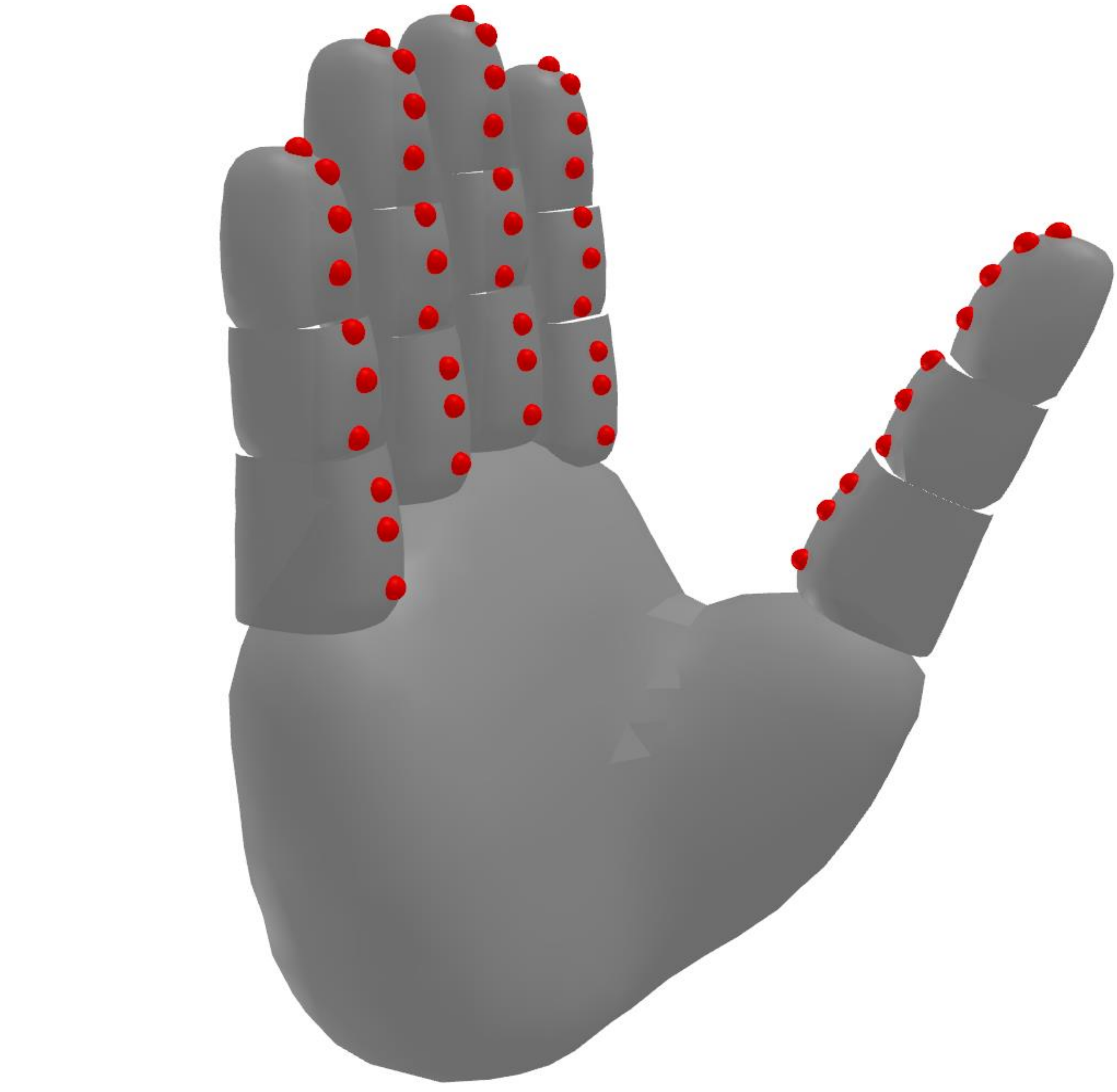}
        \label{fig:hand_contact_points}}
    \caption{(a) An example of the hand-object contact map, the red part shows the affordance map of the object, the orange part shows the contact region on the hand. (b) Simplified mesh of the HIT-DLR \uppercase\expandafter{\romannumeral2} Hand model and the sampled potential grasp points (red).}
    \label{fig:hand_object_contact}
    \vspace{-6mm}
\end{figure}

\textbf{Hand-Object Contact}. A physically plausible grasp should hold the object tightly. Intuitively, we propose to model reasonable grasp based on contacts in two folds: \textit{Which part of the object that the fingers should be in contact with?} And \textit{Which part of finger should grasp?} Specifically, given the ground-truth hand-object grasp, both object affordance points $\mathcal{O}^c$ and hand grasp vertices $\mathcal{H}^c$ can be derived from the distance between object points and hand vertices, as shown in \mbox{Fig.\ref{fig:contact_map}}. We denote the object point subset that is close enough to the hand vertices as $\mathcal{O}^c$, and the contact points on hand that is close enough to the object points as  $\mathcal{H}^c$. We formulate the contact losses as follows:
\begin{equation}
\begin{aligned}
     \mathcal{L}_{\mathcal{O}} &= \frac{1}{\left| \mathcal{O}^c \right|}\sum_{\bm{p} \in \mathcal{O}^c} {(\bm{f}(\bm{p} | \hat{\bm{V}}) - \bm{f}(\bm{p} | \bm{V}))} \\
     \mathcal{L}_{\mathcal{H}} &= \frac{1}{\left| \mathcal{H}^c \right|}\sum_{\bm{v} \in \mathcal{H}^c} {(\bm{f}(\hat{\bm{v}} | \mathcal{P}^o) - \bm{f}(\bm{v} | \mathcal{P}^o))} \\
     \mathcal{L}_{\mathcal{C}} &= \lambda_{\mathcal{O}} \cdot \mathcal{L}_{\mathcal{O}} + \lambda_{\mathcal{H}} \cdot \mathcal{L}_{\mathcal{H}}
\end{aligned}
\label{equation:contact_loss}
\end{equation}
Where $f(\bm{\cdot} | \bm{\cdot} )$ refers to Signed Distance Field (SDF), the magnitude of a point represents the distance to the surface boundary and the sign indicates whether the region is inside (-) or outside (+), $\textit{e.g.}$ $\bm{f}(\bm{p} | \bm{V})$ outputs the signed distance for point $\bm{p}$ to hand vertices set $\bm{V}$. The hand-object contact loss $\mathcal{L}_{\mathcal{C}}$ includes two terms, $\mathcal{L}_{\mathcal{O}}$ for object affordance loss and $\mathcal{L}_{\mathcal{H}}$ hand contact loss. The loss $\mathcal{L}_{\mathcal{O}}$ encourages the object affordance map to be consistent with the ground-truth affordance map, and the loss $\mathcal{L}_{\mathcal{H}}$ penalizes the difference between the predicted  hand contact region and the ground-truth. $\lambda_{\mathcal{O}}$ and $\lambda_{\mathcal{H}}$ are constants for balance.

\textbf{Interpenetration}. In order to generate realistic hand grasp, we need to take consideration of physical constraints, $\textit{i.e.}$ interpenetration between hand gripper and the target object. To alleviate hand-object intersection, we formulate interpenetration loss as follows:
\begin{equation}
\begin{aligned}
     \mathcal{L}_{\mathcal{P}} &= \sum_{\bm{p} \in \mathcal{P}^{o}}\min{(-\langle \bm{1}, f(\bm{p}|\bm{V})\rangle, 0)}
\end{aligned}
\label{equation:collision_loss}
\end{equation}
Where $\bf{1}$ is a 2D one-vector, and $\langle \cdot, \cdot \rangle$ denotes a dot product. The interpenetration loss $\mathcal{L}_{\mathcal{P}}$ actually penalize the negative sum of signed distances of the object point to the hand mesh.

Finally, the overall loss are summarized as follows:
\begin{equation}
\begin{aligned}
     \mathcal{L}_{grasp} &= \lambda_{\mathcal{KL}} \cdot \mathcal{L}_{\mathcal{KL}} + \mathcal{L}_{\mathcal{R}} + \mathcal{L}_{\mathcal{C}} + \lambda_{\mathcal{P}} \cdot \mathcal{L}_{\mathcal{P}}
\end{aligned}
\label{equation:overall_generator_loss}
\end{equation}
where $\lambda_{\mathcal{KL}}$ and $\lambda_{\mathcal{P}}$ are constants to balance the losses.
\subsection{Iterative Grasp Refinement}
Although our model can predict realistic grasps for most of the objects in our dataset, two failure cases remains, as shown in Fig.\ref{fig:diverse_grasps}: 1) Penetration with thin-walled objects, such as bowls and mugs; 2) Hand-object contact is not tight enough, causing grasp failures. Considering the above drawbacks when operating on the real robotic platform, we further refine the grasp quality to avoid collisions. To achieve this, we propose to improve the grasp quality by penalizing hand-object penetration and optimizing the grasp contact energy in our refinement module. Specifically, given the predicted grasp pose of the grasp sampler and complete point cloud, the grasp refinement module takes as input the predicted grasp $g$ and hand-object contact distance, then predicts the residual grasp $\Delta{g}$ transformation. In this way, the refinement module can work in an iterative manner: 
\begin{equation}
\begin{aligned}
     \mathcal{L}_{CE} &= \min{\sum_{\bm{p} \in \mathcal{H}^{g}}{\mathbb{I}(f(\bm{p}|\bm{V}))}}\\
     \mathbb{I}\big(f(\cdot|\cdot)) &= 
        \begin{cases}
         f(\cdot|\cdot) & \text{if } f(\cdot|\cdot) > \mathcal{T} \\
             0  & \text{otherwise}. 
         \end{cases}\\
     \mathcal{L}_{\mathcal{D}} &= \mathcal{D}(g, g^*) \\
     \mathcal{L}_{refine} &= \lambda_{CE} \cdot \mathcal{L}_{{CE}} + \lambda_{\mathcal{P}} \cdot \mathcal{L}_{\mathcal{P}} + \lambda_{\mathcal{D}} \cdot \mathcal{L}_{\mathcal{D}}
\end{aligned}
\label{equation:overall_loss}
\end{equation}
The refinement loss $\mathcal{L}_{refine}$ includes three terms: $\mathcal{L}_{CE}$ for optimizing the grasp distance contact energy by attracting potential grasp points $p \in \mathcal{H}^{g}$ to be close to the target object, which helps to produce wrap-around grasps. $\mathcal{L}_{\mathcal{P}}$ for penalizing hand-object interpenetration and  $\mathcal{L}_{\mathcal{D}}$ regularize the refined grasp $g^*$ should be close to the input grasp $g$. $\mathbb{I}$ denotes the Indicator function for judging whether the hand grasp points are close enough, $\mathcal{T}$ is the distance threshold, we set it to 2 centimeters. We define $D(g, g^*)$ as the distance measurement function, where $g$ is the predicted hand grasp of the sampler, $g^*$ is the output grasp configuration of the refinement module. We manually label $N=50$ potential grasp contact points $\mathcal{H}^{g}$ on the gripper surface, as shown in \mbox{Fig.\ref{fig:hand_contact_points}}.

Qualitative results shown in Fig.\ref{fig:iterative_refinement_examples} show our refinement algorithm can deal with inaccurate grasp postures and refine them to the pose with fewer collisions and higher quality.


\section{EXPERIMENTS}
We evaluate our model both in simulation and on a real robot platform consisting of a UR5 robotic arm equipped with a HIT-DLR \mbox{\uppercase\expandafter{\romannumeral2}} Hand.
\subsection{Implementation Details}
We sample 2048 surface points for each object during training the variational grasp generator, and these points are sampled using the Farthest Point Sampling (FPS) method. During the inference stage, 2048 points are sampled among the complete points predicted by the point completion module. The variational grasp generator and refinement network are trained for 250 and 100 epochs respectively with the learning rate set to 0.002 at start and divided by 10 when the validation error plateaus. Batch size is 512. We train our model on a RTX-3090 GPU. The dimension of the latent space is set to 4 for real robot platform evaluation. Other hyper-parameters for training our network are listed in Tab.\ref{tab:hyparameter}

\begin{table}[t]
	\caption{Hyper-parameters Setting}
	\centering
	\begin{tabular}{l|llllllll}
\hline
parameters & $\lambda_{\mathcal{KL}}$ & $\lambda_{{\theta}}$ & $\lambda_{v}$ & $\lambda_{\mathcal{O}}$ & $\lambda_{\mathcal{H}}$ & $\lambda_{\mathcal{P}}$ & $\lambda_{\mathcal{D}}$ & $\lambda_{CE}$ \\ \hline
value      & 0.1  & 0.5  & 30  & 30  & 30  &  1 & 1  & 20 \\ \hline
\end{tabular}
	\label{tab:hyparameter}
	\vspace{-6mm}
\end{table}

\subsection{Evaluate Metric} We use three quantitative metrics to evaluate our approach consistently with previous literature \cite{hasson2019learning,6dofgraspnet}.

\begin{figure}[bp]
\vspace{-4mm}
    \centering
    \includegraphics[width=0.85\linewidth]{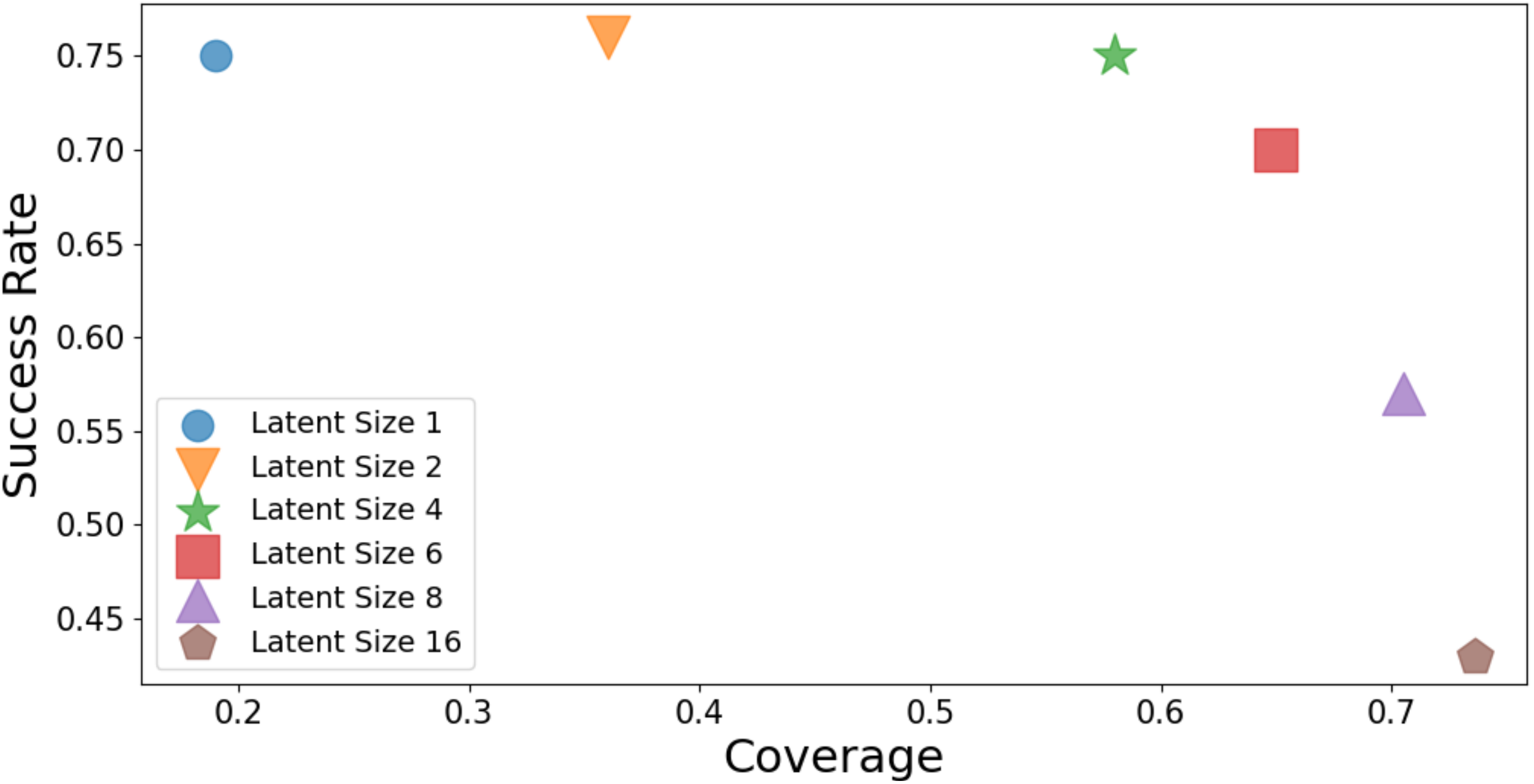}
    \vspace{-3mm}
    \caption{Scatter chart shown the influence of number of dimensions.}
    \label{fig:latent_size}
\end{figure}

\textbf{Penetration} for measurement of the penetration between the hand mesh and the target object. It consists of two terms: penetration \textbf{depth} and \textbf{volume}. We follow the implementation used in \cite{jiang2021hand}. If the hand and the object collide, the penetration depth is the maximum of the distances from hand mesh vertices to the object surface.

\textbf{Success Rate (SR)} is used to measure the stability and quality of the generated grasps, which is commonly used in grasping tasks.

\textbf{Coverage Rate} is utilized to measure the diversity of the generated grasps and measures how well the generated grasps cover the space of positive grasps $G^*$. We follow the same setting in \cite{6dofgraspnet}, that only the distance in the translation of the grasps is used for evaluating whether a grasp is covered or not. Grasp $g$ no further than 2cm away from any of the grasp $\hat{g} \in G$ will be considered as a hit.

\textbf{Time Cost} is utilized to measure the time efficiency.
\begin{figure}[t]
    \centering
    \includegraphics[width=0.9\linewidth]{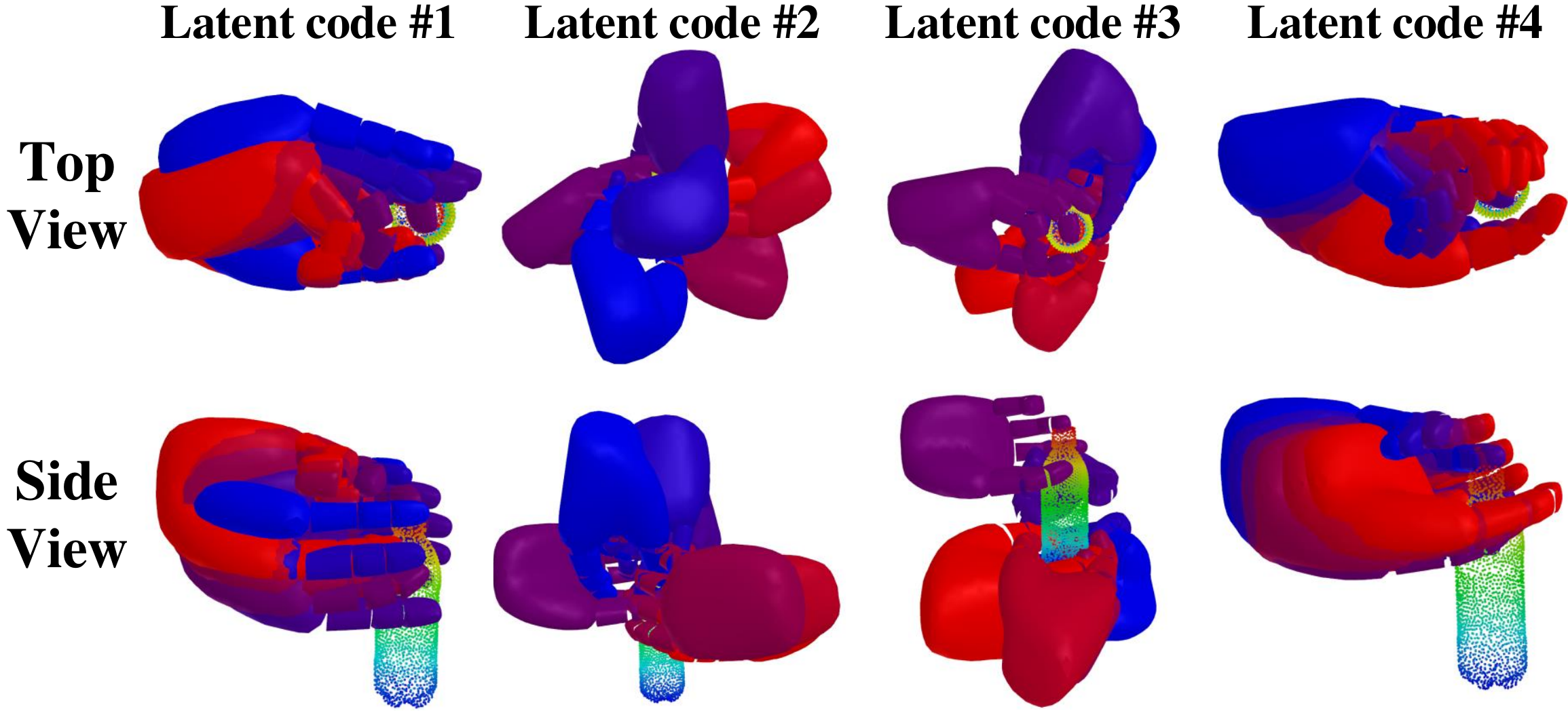}
    \vspace{-3mm}
    \caption{Qualitative results for the latent codes.}
    \label{fig:interpolated_latent_code}
    \vspace{-6mm}
\end{figure}

\subsection{Simulation Experiments} We carry out most of our experiments in the physics simulator \cite{mujoco}, since we can access complete object models which helps to evaluate the grasp sampler and the iterative refinement module. In addition, the gripper can move freely in simulation environments that we do not take motion-planning for robot arm in simulation. 58 objects from YCB dataset (seen) and 48 objects from EGAD! (unseen) are selected for comprehensive evaluation.

\begin{figure}[bp]
    \vspace{-6mm}
    \centering
    \includegraphics[width=0.85\linewidth]{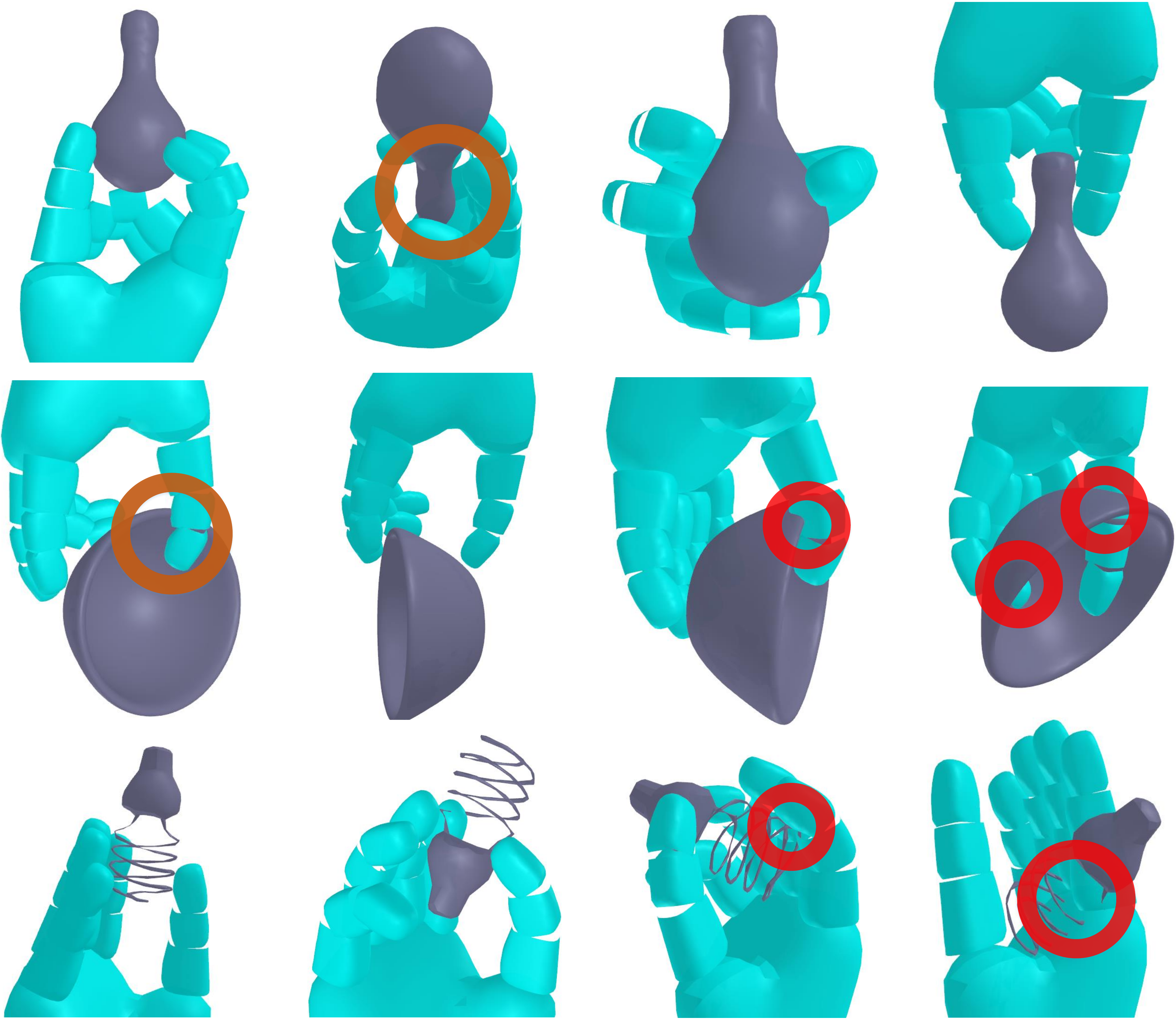}
    \vspace{-3mm}
    \caption{Qualitative results of the variational grasp generator. The red circles mark the parts with collisions and the brown circles mark the parts with no tight contacts.}
    \label{fig:diverse_grasps}
\end{figure}

\textbf{Why is VAE needed.} Firstly, our method generates, on average, a grasp in around 0.16 seconds compared to the 35+ seconds required by GraspIt!, as shown in Tab.\ref{tab:learned_vs_graspit}. VAE is 228 times faster. Secondly, given the completed object model, sample a single grasp candidate using approach-based sampling method is inefficient, diverse grasps can be obtained by choosing different approach vectors,  while graspable approach direction can be sparse for certain objects \textit{e.g.} mug and bowl. To encode the space of successful grasps efficiently, we use the VAE network.

\textbf{Dimensionality of Latent Space}. The dimensionality of the latent space is a key hyper-parameter influencing the quality of the generated grasps when training VAE. In general, a high-dimensional latent space helps to improve the capacity for reconstruction, but it can also lead to increasing the possibility of over-fitting. To study the influence of dimensionality of the latent space, we show a quantitative analysis with the increasing number of dimensions. Specifically, 1000 grasps are sampled for calculating the Success Rate and the Coverage Rate in Fig.\ref{fig:latent_size}. As shown in the figure, a dimensionality of four achieves the most appropriate balance between the Success Rate and the Coverage Rate. We choose this setting for all experiments afterwards. In \mbox{Fig.\ref{fig:diverse_grasps}}, qualitative results show that our model can generate diverse sets of grasp for a same object geometry. In Fig.\ref{fig:interpolated_latent_code}, qualitative results demonstrate what happens to the grasps when interpolating latent codes. For each dimension of the latent code, we sample 7 instances uniformly and normalize the color from blue to red, all other dimensions are set to 0. The top row shows the top view, the bottom row shows the side view. Some dimensions of the latent space encoding produce smooth interpolations between samples. For example, the first column seems to encode smooth translation in the z-axis,  the fourth column seems to encode smooth rotation in the z-axis, the second and third column both may encode approach direction and grasp position jointly.

\begin{table}[t]
\centering
\caption{Effect of Iterative Refinement. $\uparrow$: higher the better; $\downarrow$: lower the better.}
\resizebox{\linewidth}{!}{%
\begin{tabular}{c|cc|c}
                  & \multicolumn{2}{c|}{Penetration} & \multirow{2}{*}{Success Rate $\uparrow$} \\
                  & Depth (cm) $\downarrow$         & Volume (cm$^3$) $\downarrow$        &                               \\ \hline
DVGG (w/o refine) & 0.53            & 5.15             & 64.6\%                         \\
DVGG (+1 refine)  & 0.46            & 4.26             & 72.7\%                         \\
DVGG (+2 refine)  & \textbf{0.41}         & 3.87             & 74.9\%                         \\
DVGG (+3 refine)  & \textbf{0.41}            & \textbf{3.80}             & \textbf{ 75.1\% }                       
\end{tabular}%
}
\label{tab:iterative_refinement}
\vspace{-6mm}
\end{table}

\begin{table}[bp]
\vspace{-4mm}
\caption{Ablation Study on Various Loss Functions}
\centering
\label{tab:loss_function}
\begin{tabular}{lllllll}
\hline
\multicolumn{2}{l}{Loss removed}                                                & none & $L_\mathcal{C}$ & $L_\mathcal{P}$ & $L_{CE}$ & $L_\mathcal{D}$ \\ \hline
\multicolumn{2}{l}{Success Rate (\%) $\uparrow$}                                & 72.7 & 40.5            & 10.2            & 70.0     & 65.4            \\ \cline{1-2}
\multicolumn{1}{c}{\multirow{2}{*}{Penetration}} & depth (cm) $\downarrow$      & 0.46 & 0.15            & 3.62            & 0.42     & 0.52            \\ \cline{2-2}
\multicolumn{1}{c}{}                             & Volume (cm$^3$) $\downarrow$ & 4.26 & 0.92            & 80.25           & 4.06     & 5.60            \\ \hline
\end{tabular}
\end{table}

\begin{figure}[t]
    \centering
    \includegraphics[width=0.8\linewidth]{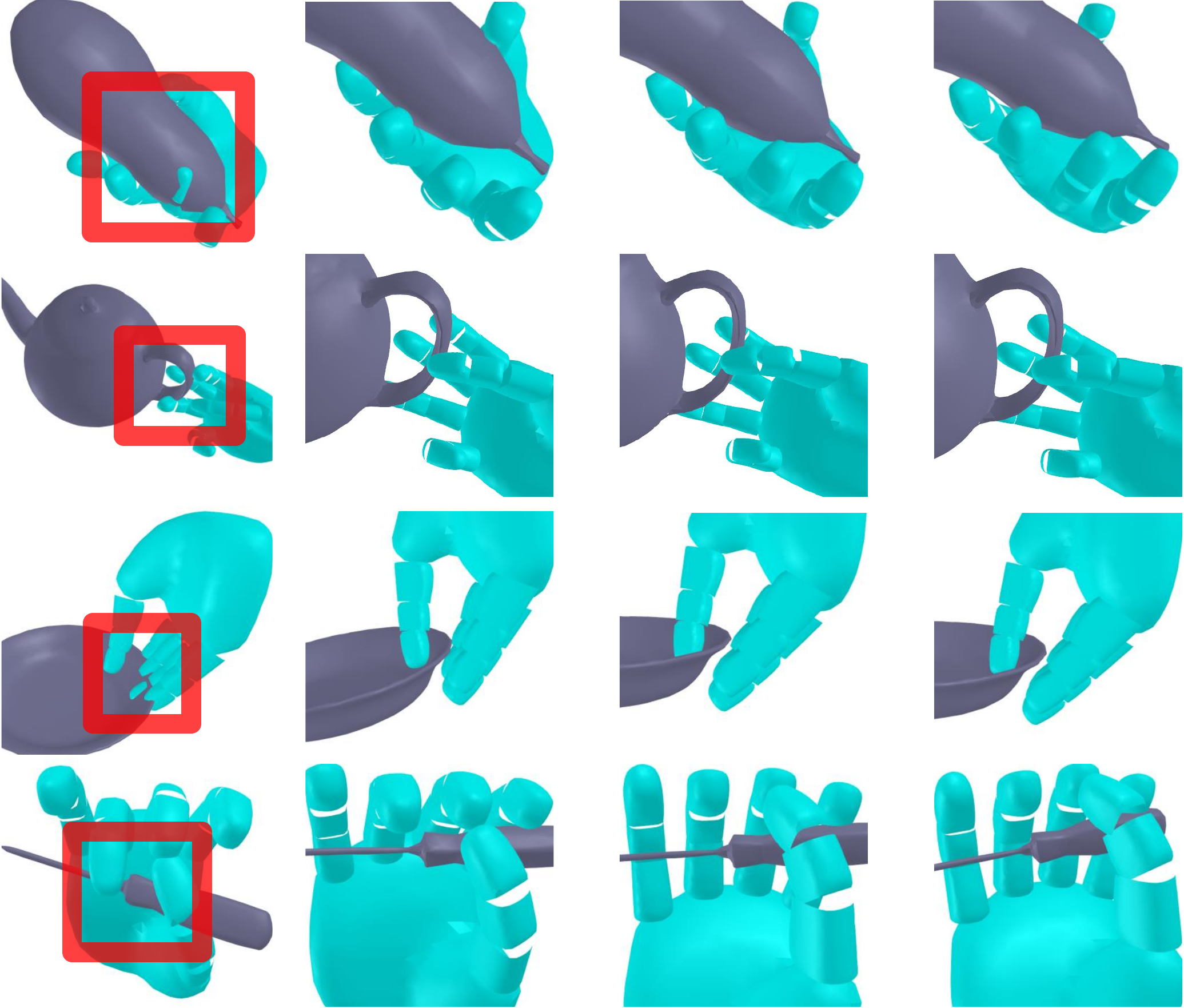}
    \vspace{-4mm}
    \caption{Cases of refinement.The first column shows coarse grasps, the right columns show grasps after 1, 2 and 3 times refinement.}
    \label{fig:iterative_refinement_examples}
    \vspace{-7mm}
\end{figure}

\textbf{Effect of Iterative Refinement}. To demonstrate the effectiveness of our proposed refinement module, we compare the performance of Success Rate and penetration and for both the grasp generator and refinement module on YCB dataset. As shown in Tab.\ref{tab:iterative_refinement}, the successful rate of the grasp configurations after refinement has 12.4\% and 16.0\% improvement respectively over the coarse grasps generated by the grasp sampler with one and two times refinements, and improvement gets saturated with higher iterations. The experimental results also demonstrate that the iterative refinement model helps in alleviating interpenetration between the hand model and the target objects.  Fig.\ref{fig:iterative_refinement_examples} shows four qualitative cases during the iterative refinement process.

\textbf{Learned Grasp Sampler Vs. GraspIt!} To illustrate the efficiency and quality of the generated grasp of our model, we follow the similar setting in \cite{multifingan} that we sample 360 grasp candidates on average in GraspIt! \cite{graspit} within 75000 steps. Most of the grasp generated by GraspIt! are of low quality, only the top 20 grasp are used for evaluation. As for our method, we randomly sample 360 grasps for computing the sampling time and 20 random grasps of them to compute other metrics. Experimental results listed in \mbox{Tab.\ref{tab:learned_vs_graspit}} show that our method outperforms baseline by a large margin in terms of grasp quality and Success Rate for both the seen and unseen objects. The results also demonstrate that: 1) the model trained with complete object point clouds outperforms the model trained with partial observed point clouds on the YCB dataset 2) the iterative refinement module can improve the grasps generated by GraspIt! as well.

\begin{figure}[bp]
    \vspace{-8mm}
    \centering
    \subfloat[]{\includegraphics[height=0.275\linewidth]{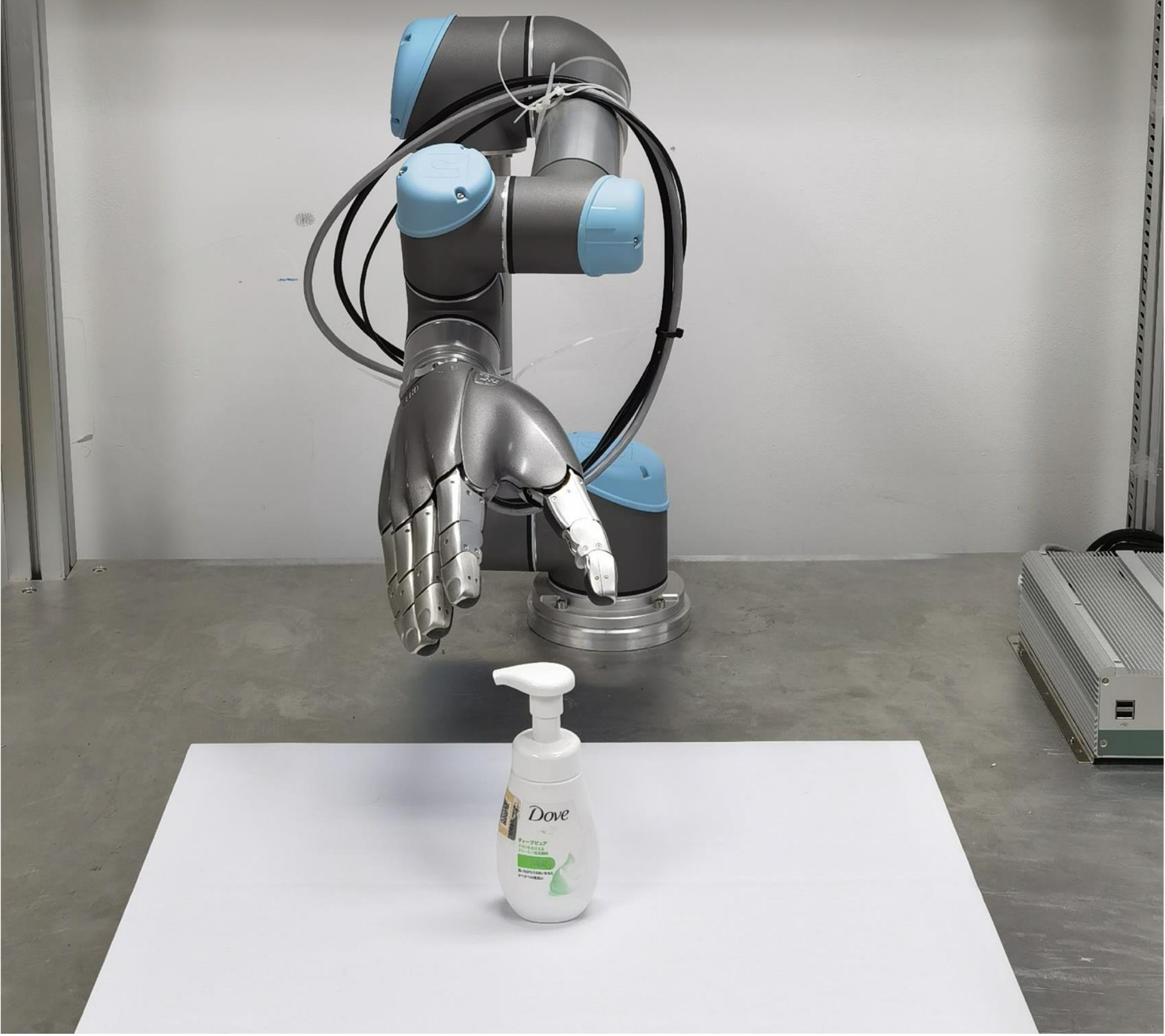}
        \label{fig:real_scene}}
    \subfloat[]{\includegraphics[height=0.275\linewidth]{figure/test_objects.pdf}
        \label{fig:test_objects}}
    \vspace{-2mm}
    \caption{Real setting of our robotic grasping experiments.
    (a) Robotic grasping experiment setup with HIT-DLR \uppercase\expandafter{\romannumeral2} Hand and UR5 robotic arm. (b) Objects used in our robotic experiments. Left part shows similar objects, right part shows novel objects.}
    \label{fig:real_scene_and_test_objects}
\end{figure}

\textbf{Effect of Various Loss}. To study the impact of various loss functions, we conduct an ablation study on these loss terms. As shown in \mbox{Tab.\ref{tab:loss_function}}, we train our model with one of these loss terms removed, and only one time refinement is applied. We show the performance of Success Rate and Penetration on the YCB dataset. As expected, the network trained without penetration loss $L_{\mathcal{P}}$ achieves the lowest performance, since the hand model often intersects the object. The model trained without hand object contact loss $L_{\mathcal{C}}$ achieves the second lowest performance in terms of Success Rate, while it intersects the object slightly. We believe this is due to the lack of contact loss, which results in most grasps being far from the object. The model trained without regularization loss $L_\mathcal{D}$  performs slightly better than  coarse grasps generated by the variational grasp generator. The network trained with no contact energy loss produces the highest Success Rate, since the contact energy loss is designed to generate tighter hand-object contact, which can not help to optimize unreasonable grasp, \textit{e.g.} grasps far away from objects and grasps with collisions.

\begin{table*}[t]
\centering
\caption{Comparison with GraspIt! in simulation. PC denotes predicted complete object points as input, PO for partial observed points, GT for ground-truth object points. * denotes adjustment with iterative refinement.}
\resizebox{0.95\textwidth}{!}{
\begin{tabular}{ccccllcccccc}
\hline
\multicolumn{2}{c}{}                                        & \multicolumn{6}{c}{YCB}                                                            & \multicolumn{4}{c}{EGAD!}                                  \\ \cline{3-12} 
\multicolumn{2}{c}{}                                        & \multicolumn{3}{c}{GraspIt!}              & \multicolumn{3}{c}{DVGG}               & \multicolumn{2}{c}{GraspIt!}    & \multicolumn{2}{c}{DVGG} \\ \hline
\multicolumn{2}{c}{Input}                                   & PC    & GT    & \multicolumn{1}{l|}{GT*} & PO   & PC*    & \multicolumn{1}{c|}{GT*} & PC    & \multicolumn{1}{c|}{GT} & PC*          & GT*         \\ \hline
\multicolumn{2}{c}{$\epsilon$-metric $\uparrow$}            & 0.028 & 0.029 & 0.034                     & 0.01 & 0.071 & 0.063                   & 0.043 & 0.044                   & 0.075       & 0.081      \\ \cline{1-2}
\multirow{2}{*}{Penetration} & depth (cm) $\downarrow$      & 0.44  & 0.25  & 0.028                     & 1.95 & 0.65  & 0.41                    & 0.31  & 0.20                    & 0.053       & 0.037       \\
                             & volume (cm$^3$) $\downarrow$ & 2.68  & 0.75  & 0.87                      & 25.3 & 4.18  & 3.84                    & 1.21  & 0.52                    & 3.90        & 3.23       \\ \cline{1-2}
\multicolumn{2}{c}{Sampling Time (sec.) $\downarrow$}       & 36.6  & 37.8  & 37.8                      & 0.16 & 0.16  & 0.16                    & 37.6  & 37.5                    & 0.16        & 0.16       \\ \cline{1-2}
\multicolumn{2}{c}{Success Rate (\%) $\uparrow$}            & 51.3  & 50.1  & 57.2                      & 27.0 & 72.4  & 74.9                    & 72.3  & 71.0                    & 75.2        & 82.2      
\end{tabular}
}
\label{tab:learned_vs_graspit}
\vspace{-3mm}
\end{table*}

\begin{figure*}[t]
\setlength{\abovecaptionskip}{0.2cm}
    \begin{center}
		\includegraphics[width=0.95\linewidth]{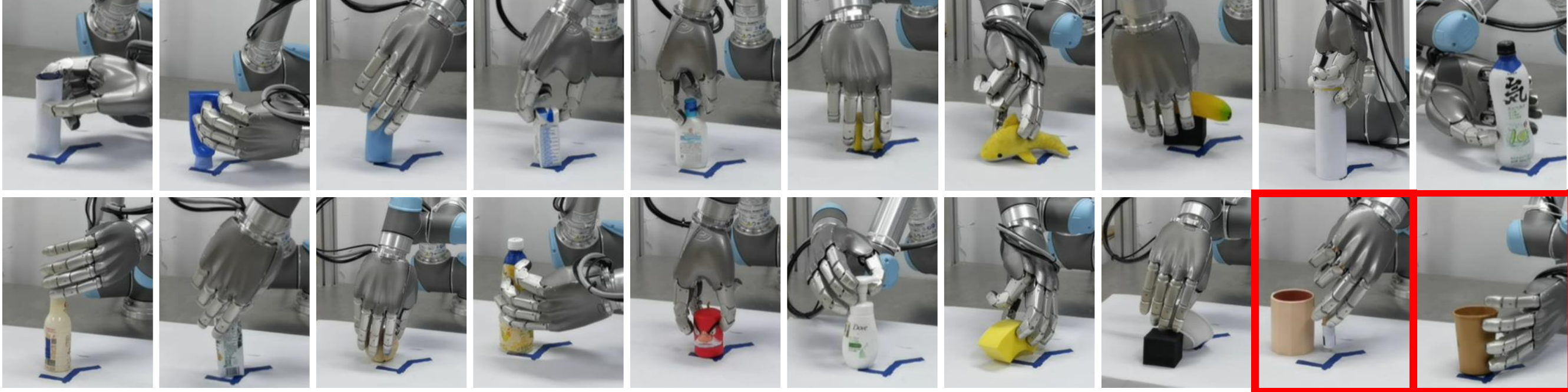}
	\end{center}
	\vspace{-5mm}
    \caption{Qualitative grasps generated by DVGG on 20 real objects. The upper row shows grasps for the similar objects, the bottom row shows grasps for the novel objects. Failure cases are shown in red boxes.}
    \label{fig:real_object_grasps}
    \vspace{-6mm}
\end{figure*}

\subsection{Robotic Experiments}
We validate the effectiveness and reliability of our method in HIT-DLR \uppercase\expandafter{\romannumeral2} Hand with a UR5 robot arm. We capture the point cloud with the Ensenso N35 camera. Objects are presented to the robot individually on top of the foam pad, as shown in Fig.\ref{fig:real_scene}. We keep the following setting in actual robot platform experiments: 1) The camera captures the scene from the backside at a 60-degree viewpoint; 2) Objects are randomly placed within a 25x25 cm square area with stable poses and their point clouds segmented by subtracting the background; 3) 10 similar and 10 novel objects are selected to evaluate the generalization ability of the proposed network, as shown in Fig.\ref{fig:test_objects}; 4) We use the point completion method \cite{vrcnet} described above; 5) Grasps cause collision with the ground will be filtered out. 6) We employ MoveIt for path planning of the UR5 robot arm. 7) We employ joint position control with extra +10 degrees for each flexion joint.
\begin{table}[bp]
    \vspace{-6mm}
    \centering
    \caption{Results of Robotic Platform Experiments}
    \resizebox{0.875\linewidth}{!}{%
        \begin{tabular}{l|c|c}
        \multicolumn{1}{c|}{} & \multicolumn{1}{c|}{Similar objects} & \multicolumn{1}{c}{Novel objects} \\
        \multicolumn{1}{c|}{} & SR (Avg)$\uparrow$ & SR (Avg)$\uparrow$        \\ \hline
        GraspIt!\cite{graspit}          & 48.7\% & 46.0\% \\
        PointNetGPD\cite{pointnetgpd}       & 52.3\% & 48.3\% \\
        DVGG (w/o refine) & 67.3\% & 57.3\% \\
        DVGG (+3 refine)  & \textbf{74.7\%} & \textbf{70.7\%}
        \end{tabular}%
        }
    \label{tab:robot_experiments}
\end{table}

We compare \textbf{DVGG} to GraspIt! and PointNetGPD \cite{pointnetgpd}. GraspIt! requires triangle-mesh based object model, we follow the poisson surface reconstruction algorithm to complete the target surface. Since PointNetGPD is proposed to classify parallel-jaw grasps, we take the following modifications: 1) We use GraspIt! to sample grasps. 2) Given the sampled grasps, we crop the point clouds within fixed radius to train the classification model proposed by [2].


The experimental procedure is as follows: 1) We generate 15 grasps per object for each algorithm; 2) Only physically reachable grasps will be executed. For GraspIt!, we adopt the widely used $\epsilon$-metric to select grasps. As for PointnetGPD and our method, the first physical reachable grasp is executed. 3) A grasp will be classified as a successful grasp if the robot hand can grasp the target object and lift it to the predefined position without dropping it during translation. 

As shown in Tab.\ref{tab:robot_experiments}, our method outperforms baseline methods by a large margin, which demonstrates the superiority of our method. Qualitative results shown in Fig.\ref{fig:real_object_grasps} demonstrate that DVGG is able to generate diverse sets of grasps in terms of  posture and grasping area. 

In fact, DVGG also produced some failed grasps. Two examples are shown in \mbox{Fig.\ref{fig:real_object_grasps}}. The main reason causing grasp failure is that unstable grasps require large frictional forces to lift the target object. As shown in the second example from the right in the bottom row, the failure grasp is caused by the low friction between the hand and the target object. Another reason for grasp failure is collision between the hand and the target object. The failure shown in the rightmost image of the bottom row is caused by displacement of the object due to slight collision during hand approaching.

\section{CONCLUSIONS}
In this paper, we propose DVGG for generating diverse sets of grasps for high-DOF anthropomorphic robotic hand. Our method focuses on hand-object interaction constrain, which helps to estimate physical plausible grasps. Meanwhile, we build a large-scale synthetic grasping dataset with 300 objects with various shape. Experiments show that our model trained on the synthetic dataset performs well in real-world scenarios and outperforms the baseline by a large margin.


\bibliographystyle{IEEEtran}
\bibliography{reference}

\end{document}